\documentclass{article}
\usepackage{spconf,amsmath,graphicx,subfigure}
\usepackage{booktabs} 

\title{Submodular Video Object Proposal Selection for Semantic Object Segmentation}

\name{Tinghuai Wang}
\address{Nokia Labs, \\	Nokia Technologies, Finland}

\begin{document}
%
\maketitle
\begin{abstract}
Learning a data-driven spatio-temporal semantic representation of the objects is the key to 
coherent and consistent labelling in video. 
This paper proposes to achieve semantic video object segmentation by learning a data-driven representation 
which captures the synergy of multiple instances from continuous frames.
To prune the noisy detections, we exploit the rich information among multiple instances and select the discriminative and representative 
subset.  This selection process is formulated as a facility location problem solved by maximising a submodular function.
Our method 
 retrieves the longer term contextual dependencies which
 underpins a robust semantic video object segmentation algorithm. 
 We present extensive experiments on a challenging dataset that demonstrate the superior performance
 of our approach compared with the state-of-the-art methods.

\end{abstract}
\section{Introduction}
\label{sec:intro}
 The proliferation of user-uploaded videos which are frequently associated with semantic tags provides a vast resource for computer vision research. These semantic tags, albeit not spatially or temporally located in the video, suggest visual concepts appearing in the video. This social trend has led to an increasing interest in exploring the idea of segmenting video objects with weak supervision or labels. 

Hartmann {\em et al.} \cite{HartmannGHTKMVERS12} firstly formulated the problem as learning weakly supervised classifiers for a set of independent spatio-temporal segments.
Tang {\em et al.} \cite{TangSY013} learned discrimative model by leveraging labelled positive videos and a large collection of negative examples based on distance matrix. Liu {\em et al.} \cite{LiuTSRCB14} extended the traditional binary classifition problem to multi-class and proposed nearest neighbor-based label transfer algorithm which encourages smoothness between regions that are spatio-temporally adjacent and similar in appearance. Zhang {\em et al.} \cite{ZhangCLWX15} utilized pre-trained object detector to generate a set of detections and then pruned noisy detections and regions by preserving spatio-temporal constraints. 


Graphical models have emerged as powerful tools in computer vision
\cite{wang2010multi,qi20173d,wang2017submodular,wang2019zero,wang2019graph,yang2021learning}, offering a versatile framework for representing and analyzing complex visual data.
These approaches leverage the inherent structure and relationships within images
\cite{wang2015robust,tinghuai2016method,xing2021learning} and videos
\cite{wang2010video,wang2014wide,chen2020fine}, enabling more sophisticated and context-aware analysis. By representing visual elements as nodes and their interactions as edges, graphical models can capture spatial, temporal, and semantic dependencies crucial for various tasks such as visual information retrieval \cite{hu2013markov}, stylization \cite{WangCSCG10,wang2011stylized,wang2013learnable}, object detection \cite{WangW16,tinghuai2016apparatus,zhao2021graphfpn}, scene understanding \cite{WangW14,wang2017cross,wang2020spectral,tinghuai2020watermark,deng2021generative}, and image or video segmentation
\cite{wang2015weakly,wang2016semi,wang2016primary,tinghuai2017method,tinghuai2018method1,ZhuWAK19,zhu2019cross,tinghuai2020semantic,lu2020video,wang2021end}.

In contrast to previous works, we propose to learn a class-specific representation which captures the synergy of multiple instances  from continuous frames. To prune the noisy detections, we exploit the rich information among multiple instances and select the discriminative and representative 
subset. In this framework, our algorithm is able to bridge the gap between image classification and video object segmentation, leveraging the ample pre-trained image recognition models rather than strongly-trained object detectors. 


\section{Object Discovery}
\label{sec:discovery}

\begin{figure}[t!]
	\centering
	\includegraphics[width=0.99\linewidth]{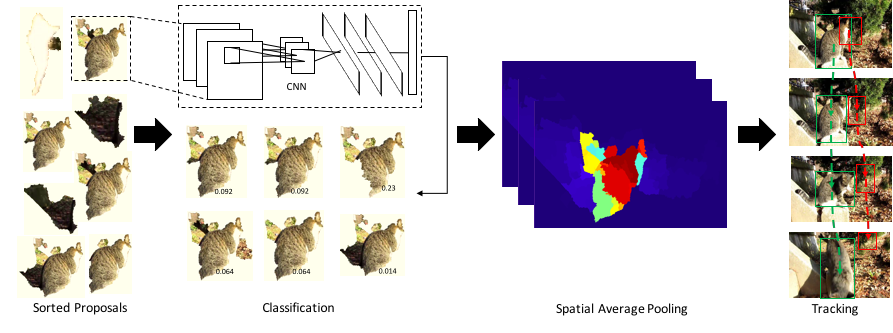}
	\caption{An illustration of the proposed object discovery strategy.}\label{fig:discovery}
\end{figure}

Semantic object segmentation requires not only localising objects of interest within a video, but also assigning class label for pixels belonging to the objects. One potential challenge of using pre-trained image recognition model to detect objects is that any regions containing the object or even part of the object, might be ``correctly'' recognised, which results in a large search space to accurately localise the object. To narrow down the search of targeted objects, we adopt category-independent bottom-up object proposals  \cite{EndresH10}. The proposed object discovery strategy is illustrated in Fig. \ref{fig:discovery}, in which the key steps are detailed in the following sections. 

\subsection{Proposal Scoring and Classification}
We combine the objectness score associated with each proposal from Endres and Hoiem \cite{EndresH10}  and motion information as a context cue to characterise video objects. We follow Papazoglou and Ferrari \cite{Papazoglou2013}  which roughly produces a binary map indicating whether each pixel is inside the motion boundary after compensating camera motion. After acquiring the motion cues, we score each proposal $r_i$ by both appearance and context,  $s_{r_i} = \mathcal{A}(r_i)+ \mathcal{C}(r_i)  \label{eq:objness1}$, 
where $\mathcal{A}(r_i)$ indicates region level appearance score computed using \cite{EndresH10} and $\mathcal{C}(r_i)$ represents the motion score of region $r_i$ which is defined as $\mathcal{C}(r_i) = \mathrm{Avg}(M^t(r_i))\cdot \mathrm{Sum}(M^t(r_i))$
where $\mathrm{Avg}(M^t(r_i))$ and $\mathrm{Sum}(M^t(r_i))$ compute the average and total amount of motion cues \cite{Papazoglou2013} included by proposal $r_i$ on frame $t$ respectively. Note that appearance, contextual and combined scores are normalised. 

To classify each scored region proposal, we firstly warp all pixels in a  bounding box around it to the required size compatible with the CNN (VGG-16 net \cite{vggnet} requires inputs of a fixed $224\times 224$ pixel size), regardless its original size or shape. Prior to warping, we expand the tight bounding box by a certain number of pixels (10 in our system) around the original box, which was proven effective in the task of using image classifier for object detection task \cite{girshick2014}. 

After the classification, we collect the confidence of regions with respect to the specific classes associated with the video and form a set of scored regions. For each proposal $r_i$, we rescore it by multiplying its score and classification confidence, which is denoted by $\tilde{s}_{r_i} = s_{r_i} \cdot c_{r_i}$. 

To ensemble the multiple confidences of region proposals in each frame, we adopt a simple spatial average pooling strategy to aggregate the region-wise confidence as well as their spatial extent. For each proposal $r_i$, we generate confidence map $\mathcal{C}_{r_i}$ of the size of image frame, which is composited as the binary map of current region proposal multiplied by its confidence $c_{r_i}$. We perform an average pooling over the confidence maps of all the proposals on each frame to compute a aggregated confidence map,
\begin{equation} 
\label{eq:conf1}
\mathcal{C}^t = \frac{\sum _{r_i \in \mathcal{R}^t} \mathcal{C}_{r_i}}{\sum _{r_i \in \mathcal{R}^t} c_{r_i}} 
\end{equation}
where $\sum _{r_i \in \mathcal{R}^t} \mathcal{C}_{r_i}$ performs element-wise operation and $\mathcal{R}^t$ represents the set of candidate proposals from frame $t$. 

\begin{figure}[t!]
	\centering
	\includegraphics[width=0.99\linewidth]{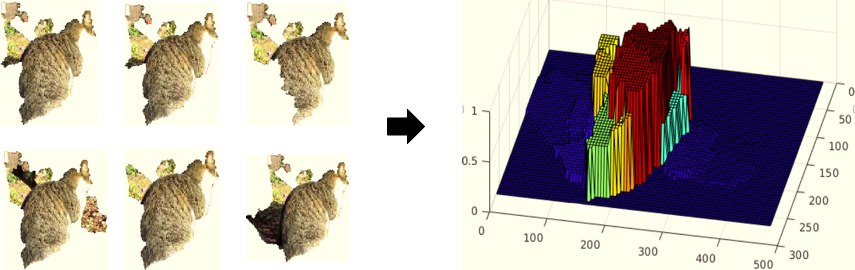}
	\caption{An illustration of the weighted spatial average pooling strategy.}\label{fig:pooling}
\end{figure}

The resulted confidence map $\mathcal{C}^t$ aggregates not only the region-wise confidence but also their spatial extent. The key 
insight is that good proposals coincide with each other in the spatial domain and their contribution to the final confidence map are proportional to their region-wise confidence. An illustration of the weighted spatial average pooling is shown in Fig. \ref{fig:pooling}.

\subsection{ Tracking for Proposal Mining}
\label{sec:tracking}
Based on the computed confidence map in Eq. \ref{eq:conf1}, we generate a new set of region proposals in a process analogous to the watershed algorithm, i.e., we gradually increase the threshold in defining binary maps from confidence map $\mathcal{C}^t$. This approach effectively exploit the topology structure of the confidence map. The disconnected regions thresholded at each level form the new proposals. The confidence associated with these new region proposals $\mathcal{P}$ are computed by averaging the confidence values enclosed by each region. 

Due to the 2D projections, it is not possible to learn a complete representation of the object in one frame, whereas multiple image frames encompassing the same object or part of the object provide more comprehensive information. Video data naturally encodes the rich information of the objects of interest. 

\begin{figure}[t!]
	\centering
	\includegraphics[width=0.99\linewidth]{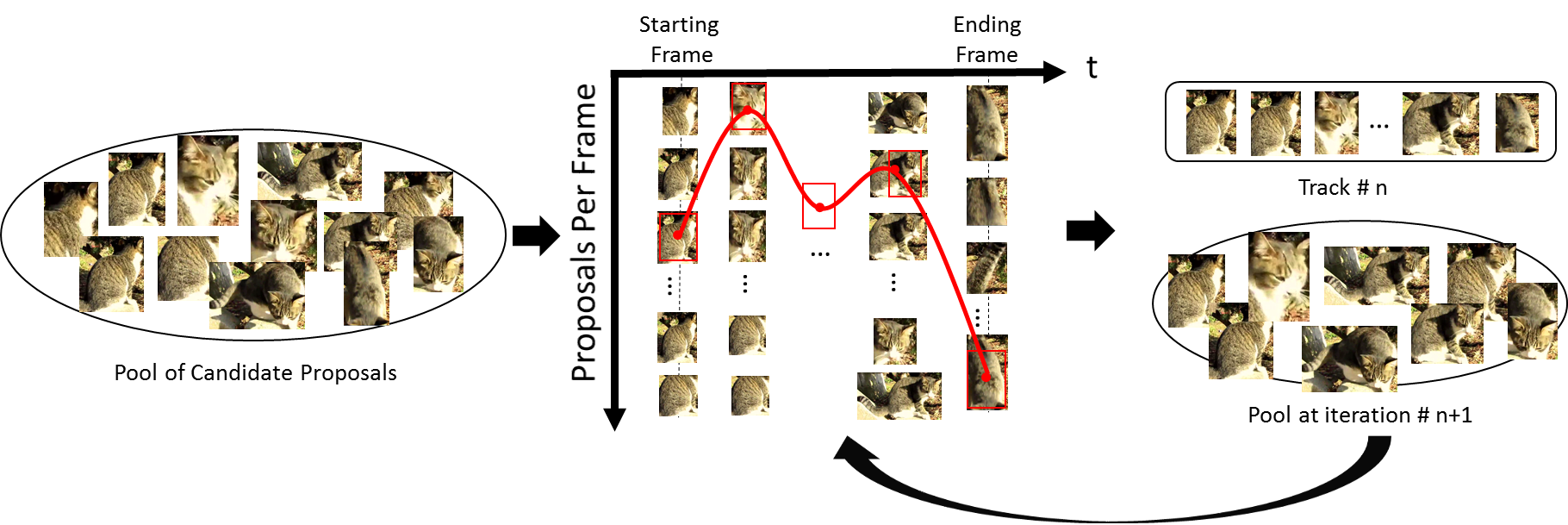} 
	\caption{Iterative tracking to eliminate spurious detections and extract consistent proposals.  \label{fig:tracks}}
\end{figure}

We propose an iterative tracking and eliminating approach to achieve these goals, as illustrated in Fig. \ref{fig:tracks}. Proposals from all frames form a pool of candidates. Each iteration 
starts by randomly selecting a proposal on the earliest frame in the pool of candidate proposals, and it is tracked \cite{danelljan2015learning} until the last frame of the sequence. Any proposals whose bounding boxes with a substantial 
intersection-over-union (IoU) overlap (0.5 in the system) with the tracked bounding box are chosen to form a track and removed from the pool. 
This process iterates until the candidate pool is empty, and forms a set of tracks $\mathcal{T}$ with single-frame tracks discarded.

\section{Submodular Track Selection}
\label{sec:LRR}

Given cohorts of noisy region proposals detected from video, we aim to select discriminative tracks for underpinning object segmentation. These detected tracks of region proposals may comprise false positive detections. These false positive detections may corrupt the representation learning object segmentation. Yet, pruning these sporadic false positive detections is not straightforward due to the limited recognition power of image classifier. Nonetheless, exploiting the similarities among tracks within the same category enables recovering the discriminative subset of tracks. 

For all the tracks $\mathcal{T}$ from each category, we construct a graph $\mathcal{G}_T=(\mathcal{V},\mathcal{E})$, where each node is a track $T_i \in \mathcal{T}$ and the edges model the pairwise relations. We aim to discover a subset of tracks  $\mathcal{D}$ of $\mathcal{T}$  by iteratively selecting elements of $\mathcal{T}$ into $\mathcal{D}$.

In order to obtain the discriminative tracks, we model the selection process as a facility location problem \cite{lazic2009floss,tsai2016semantic}, which can be formulated by a facility location term to compute the shared similarities and a discriminative term to preserve the discriminativity power of the selected tracks. 

The facility location term is defined as, 
\begin{equation} 
\mathcal{H}(\mathcal{D}) =  \sum_{i\in\mathcal{V}} \max_{j\in\mathcal{D}} w_{ij} - \sum_{i\in\mathcal{D}}  \phi_i 
\end{equation}
where $w_{ij}$ is the pairwise relation between a client node $v_i$ and a potential facility node $v_j$, and $\phi_i$ is the cost to open a facility fixed to a constant $\delta$. We define $w_{ij}$ as the similarity between tracks $T_i$ and $T_j$ to encourage $v_i$ to well represent or be similar to its clients, so that the final selected set $\mathcal{D}$ is representative.  The weight $w_{ij}$ is computed as:
\begin{equation} 
w_{ij} = <F_i, F_j>,
\end{equation}
where $F_i$ denotes the feature vector of track $T_i$, which is computed by averaging L2-normalised \emph{fc6} feature vectors from its constituent object proposals. 

To enforce a class-purity constraint on the selected tracks while preserving the submodularity, a discriminative term is defined as:
\begin{equation} 
\mathcal{P} (\mathcal{D}) = \lambda \sum_{i\in\mathcal{D}} \Phi(i),
\end{equation}
where $ \lambda$ is a weight to balance the two terms, and $\Phi(i)$ denotes the averaged confidence of all constituent object proposals of track $T_i$. 

We combine the facility location term and the discriminative term into an objective function:
\begin{align} 
\label{eq:submodular}
\max_{\mathcal{D}} \mathcal{E}(\mathcal{D})  = \max_{\mathcal{D}} \mathcal{H}(\mathcal{D})  + \mathcal{P} (\mathcal{D}),\\\nonumber
s.t. ~\mathcal{D} \subseteq \mathcal{T}  \subseteq  \mathcal{V}, N_{\mathcal{D}} \leq K,
\end{align}
where $N_{\mathcal{D}}$ denotes the number of open facilities and $K$ is the number of nodes. We optimise (\ref{eq:submodular}) using 
a greedy algorithm similar to \cite{zhu2014submodular}. 
We then
perform an average pooling over the score maps of all the selected tracks of proposals to compute confidence maps with respect to each category.

\section{Object Segmentation}
\label{sec:segmentation}

We formulate video object segmentation as a superpixel-labelling problem of assigning each superpixel with a label which represents background  or the object class  respectively. We define a space-time
graph $\mathcal{G}_s=(\mathcal{V},\mathcal{E})$ by connecting frames temporally with optical flow displacement. To achieve the efficiency and local smoothness of inference, each of the nodes in this graph is a superpixel as opposed to
a pixel, and edges are set to be the spatial neighbours within the same frame and the temporal neighbours in adjacent frames connected by least one motion vector. 


We define the energy function that minimises to achieve the optimal labelling:
\begin{small}
	\begin{equation} 
	\label{eq:graphcut}
	E(x) = \sum_{i\in \mathcal{V}} (\psi _{i}^{c}(x_i) + \lambda_{o} \psi _{i}^{o}(x_i)) + \lambda_{p} \sum_{i\in \mathcal{V}, j\in N_{i}} \psi _{i,j}(x_i,x_j) 
	\end{equation}
\end{small}
where $N_{i}$ is the set of superpixels adjacent to superpixel $s_i$ spatially and temporally in the graph respectively; $\lambda_{o}$ and $\lambda_{p}$ are parameters; $\psi _{i}^{c}(x_i)$ indicates the 
colour based unary potential and $\psi _{i}^{o}(x_i)$ is the  unary potential of
semantic object confidence which measures how likely the superpixel to be labelled by $x_i$ given the semantic confidence map;  $\psi _{i,j}(x_i,x_j)$ is the 
 pairwise potential. 

Colour unary potential is defined similar to \cite{RotherKB04}, $\psi _{i}^{c}(x_i) =  - \text{log} U_{i}^{c}(x_i)$, 
where $U_{i}^{c}(\cdot)$ is the color likelihood from colour model. Two GMM colour models are estimated over the RGB values of superpixels, for objects and background respectively, by sampling the superpixel colors according to the semantic confidence map.

Semantic unary potential is defined to evaluate how likely the superpixel to be labelled by $x_i$ given the semantic confidence map $c_i^t$ as $\psi _{i}^{o}(x_i) =  - \text{log} U_{i}^{o}(x_i)$,
where $U_{i}^{o}(\cdot)$ is the semantic likelihood, i.e., for a foreground labelling $U_{i}^{o} = c_i^t$ and $1-c_i^t$ otherwise.

We define the pairwise potentials to encourage both spatial and temporal smoothness of labelling while preserving discontinuity in the data,
\begin{equation} 
\psi _{i,j}(x_i,x_j) = [x_i \neq x_j] \text{exp}(-d^{c}(s_i,s_j))\nonumber
\end{equation}
where $[\cdot]$ denotes the indicator function. The function $d^{c}(s_i,s_j)$ computes the colour distance between spatially neighbouring superpixels $s_i$ and $s_j$ as $d^{c}(s_i,s_j) = \frac{||c_i-c_j||^2}{2<||c_i-c_j||^2>} $, where $||c_i-c_j||^2$ is the squared Euclidean distance between two adjacent superpixels in RGB colorspace. We adopt  alpha expansion  \cite{BoykovVZ01} to minimize Eq. \ref{eq:graphcut} and the resulting label assignment gives the semantic object segmentation of the video. 

\section{Experiments}
\label{sec:evaluation}
In order to evaluate the performance of semantic video object segmentation, many motion segmentation or figure-ground
segmentation datasets such as SegTrack~\cite{TsaiFNR12},~FBMS \cite{ochs2014segmentation} and~DAVIS \cite{Perazzi2016} are
not suitable due to either the ambiguous object annotation in ground-truth (one label for all foreground moving objects or
no annotation for static objects) or the insufficient number of videos/frames per object class. 
Therefore, we evaluate our method on a large scale video dataset YouTube-Objects \cite{PrestLCSF12}. YouTube-Objects consists of videos from $10$ object classes with pixel-level ground truth for every $10$ frames of $126$ videos (totally more than $20,000$ frames) provided by \cite{jain2014supervoxel} which is suitable for evaluating semantic object segmentation.  These videos are very challenging and completely unconstrained, with objects of similar colour to the background, fast motion, non-rigid deformations, and fast camera motion. 

We measure the segmentation performance using the standard IoU overlap as accuracy metric. 
We compare our approach SPS with $6$ state-of-the-art automatic approaches on this dataset, including two motion driven segmentation approaches (LTT \cite{BroxM10} and FOS \cite{Papazoglou2013}), three weakly supervised semantic segmentation approaches (LDW \cite{PrestLCSF12}, DSW \cite{TangSY013} and  SSW \cite{ZhangCLWX15}), and one object-proposal based approach (KOS \cite{LeeKG11}).

\begin{small}
\begin{table}[t!]
	\caption{Intersection-over-union overlap accuracies on YouTube-Objects Dataset}
	\centering 
	\small
	\resizebox{\columnwidth}{!}{
	\begin{tabular}{lccccccc} 
		\toprule
		& LTT \cite{BroxM10} & KOS \cite{LeeKG11} & LDW \cite{PrestLCSF12} & FOS \cite{Papazoglou2013} & DSW \cite{TangSY013}  &  SSW \cite{ZhangCLWX15}  & SPS\\
		\midrule
		Plane & 0.539 & NA  & 0.517  & 0.674  & 0.178  & \underline{\textbf{0.758}}        & 0.703\\
		Bird    & 0.196 & NA  & 0.175  & 0.625  & 0.198  & 0.608      & \underline{\textbf{0.631}} \\
		Boat   & 0.382 & NA  & 0.344  & 0.378  & 0.225  & 0.437       & \underline{\textbf{0.659}} \\
		Car     & 0.378 & NA  & 0.347  & 0.670  & 0.383  &  \underline{\textbf{0.711}}        & 0.625\\
		Cat	  & 0.322 & NA  & 0.223  & 0.435  & 0.236  & 0.465       &  \underline{\textbf{0.497}} \\
		Cow   & 0.218 & NA  & 0.179  & 0.327   & 0.268  & 0.546        &  \underline{\textbf{0.701}} \\
		Dog    & 0.270 & NA  & 0.135  & 0.489   & 0.237  & \underline{\textbf{0.555}}       & 0.532 \\
		Horse & 0.347 & NA  & 0.267  & 0.313   & 0.140  & \underline{\textbf{0.549}}         &  0.524\\
		Mbike & 0.454 & NA  & 0.412  & 0.331   & 0.125  & 0.424     	&  \underline{\textbf{0.554}} \\
		Train   & 0.375 & NA & 0.250  & \underline{\textbf{0.434}}    & 0.404  & 0.358	 	&  0.411 \\
		\midrule
		\shortstack{Cls.\\ Avg.}  &0.348 &0.28 &0.285 &0.468   & 0.239  & 0.541     	& \underline{\textbf{0.584}} \\
		\bottomrule
	\end{tabular} \label{tbl:yto-result} 
}
\end{table}
\end{small}

As shown in Table \ref{tbl:yto-result}, our method  surpasses the competing methods in  $5$ out of $10$ classes, with gains up to $4.3\%$ in category average accuracy over the best competing method SSW \cite{ZhangCLWX15}. This is remarkable considering that SSW employed strongly-supervised deformable part models (DPM) as object detector while our approach only leverages image recognition model which lacks the capability of localising objects. SSW  outperforms our method on  \emph{Car}, \emph{Dog} and \emph{Horse}, otherwise exhibiting varying performance across the categories --- higher accuracy on fast moving objects but lower accuracy on more flexible objects \emph{Bird}, \emph{Cat} and slowly moving object \emph{Cow} and \emph{Train}. We owe it to that, though based on object detection, SSW prunes noisy detections and regions by enforcing spatio-temporal constraints, rather than learning an adapted data-driven representation  in our approach. It is also worth highlighting  the improvement in classes, e.g.,  \emph{Cow},  where the existing methods normally fail or underperform due to the heavy reliance on motion information. The main challenge of the \emph{Cow} videos is that cows very frequently stand still or move with mild motion, which the existing approaches might fail to capture whereas our proposed method excels by learning an object-specific representation in low-rank constrained deep feature space to retrieve the long term dependencies in the spatial-temporal domain. Our method doubles or triples the accuracy of another weakly supervised method DSW \cite{TangSY013}  except a weaker strength on  \emph{Train} category. This is probably owing to that DSW uses a large number of similar training videos which may capture objects in rare view.  Motion driven method  FOS \cite{Papazoglou2013} can better distinguish rigid moving foreground objects on videos exhibiting relatively clean backgrounds, such as \emph{Plane}, \emph{Car} and  \emph{Train}.  Note that, FOS is a Figure-Ground segmentation method which segments all ``foreground'' moving objects without assigning semantic labels. 

\begin{figure}[t!]
	\centering
	\subfigure[Aeroplane]{
		\includegraphics[width=0.17\linewidth]{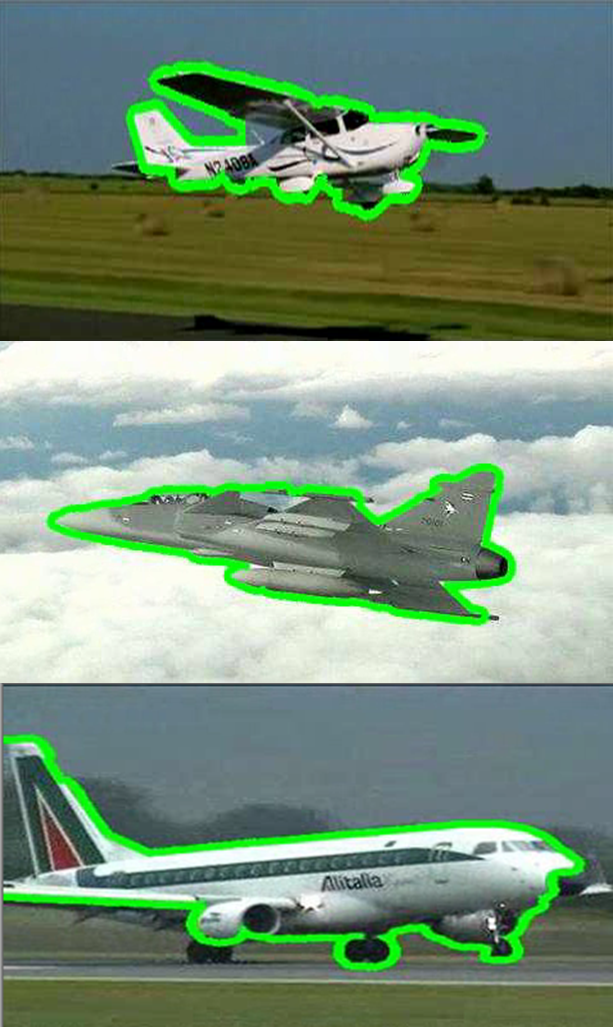}  
	}
	\subfigure[Bird]{
		\includegraphics[width=0.17\linewidth]{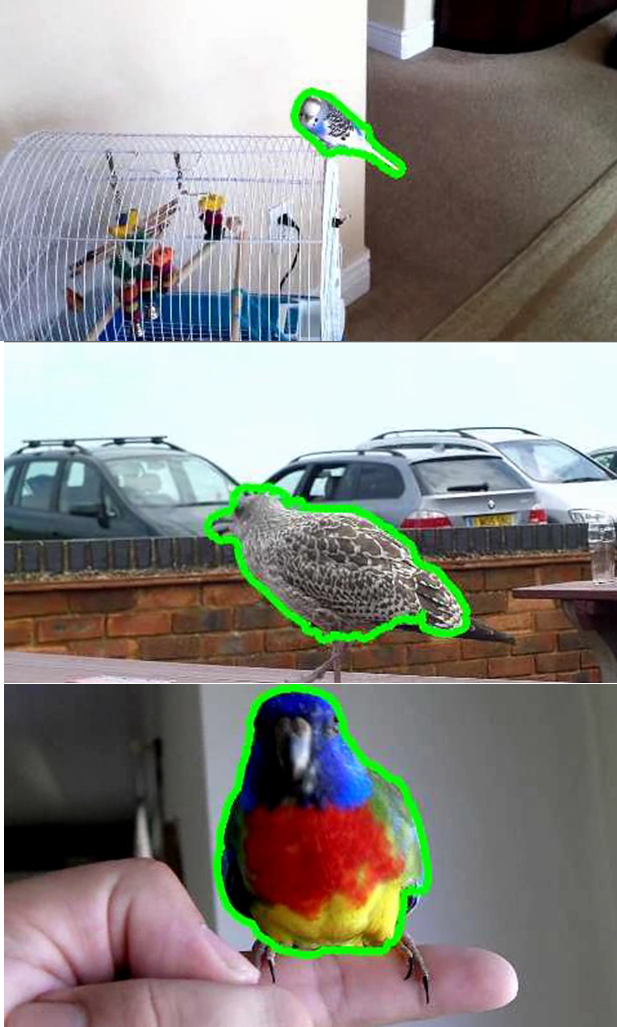}  
	}
	\subfigure[Boat]{
		\includegraphics[width=0.17\linewidth]{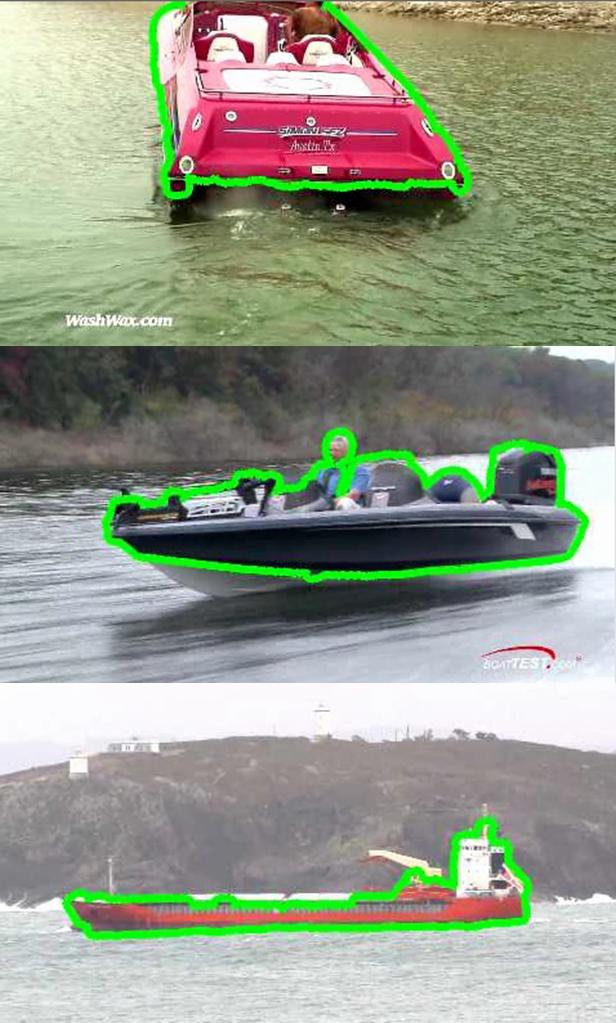}  
	}
	\subfigure[Car]{
		\includegraphics[width=0.17\linewidth]{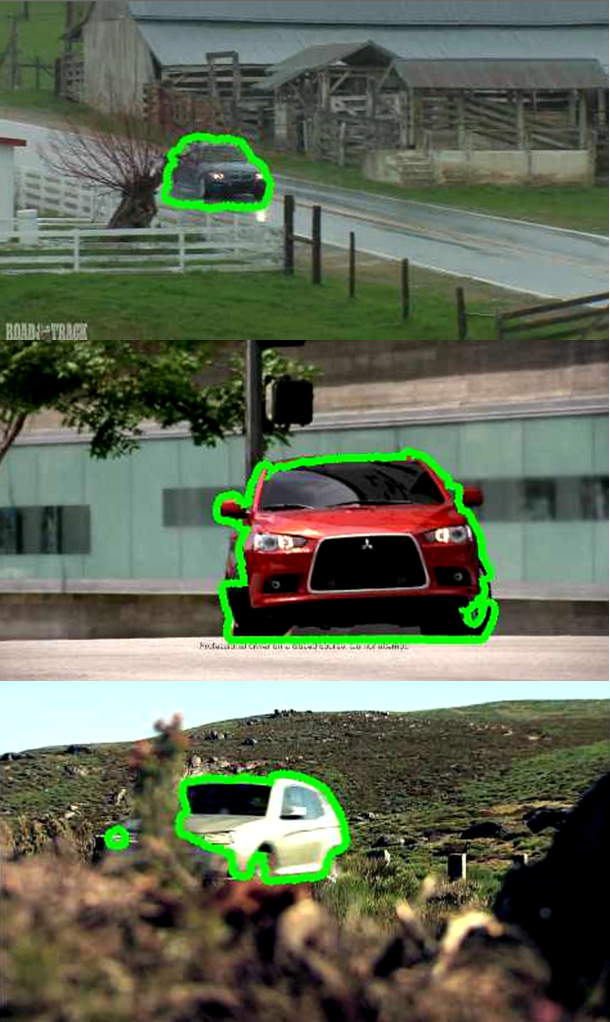}  
	}
	\subfigure[Cat]{
		\includegraphics[width=0.17\linewidth]{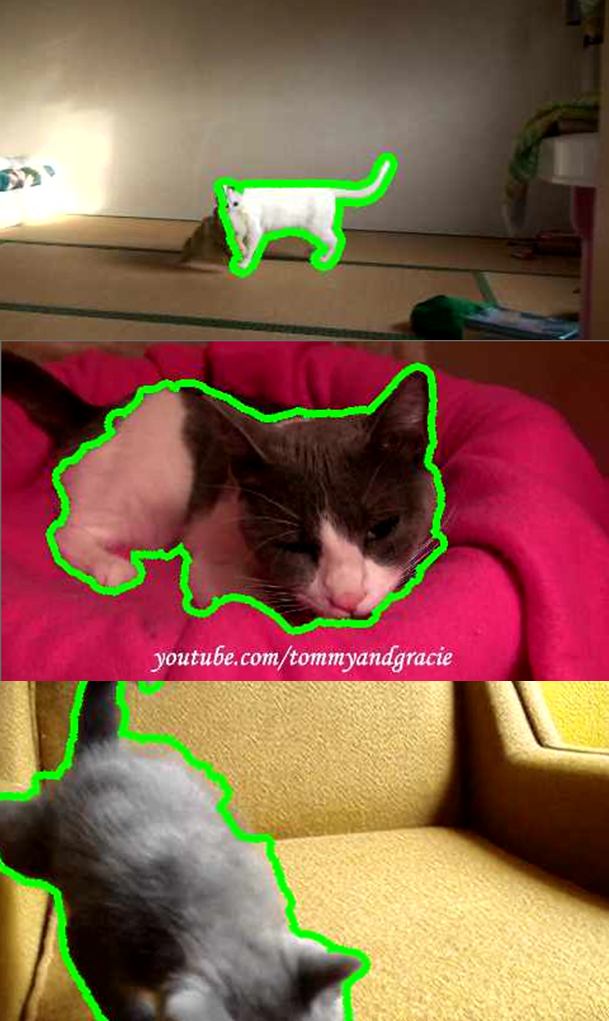}  
	}
	\subfigure[Cow]{
		\includegraphics[width=0.17\linewidth]{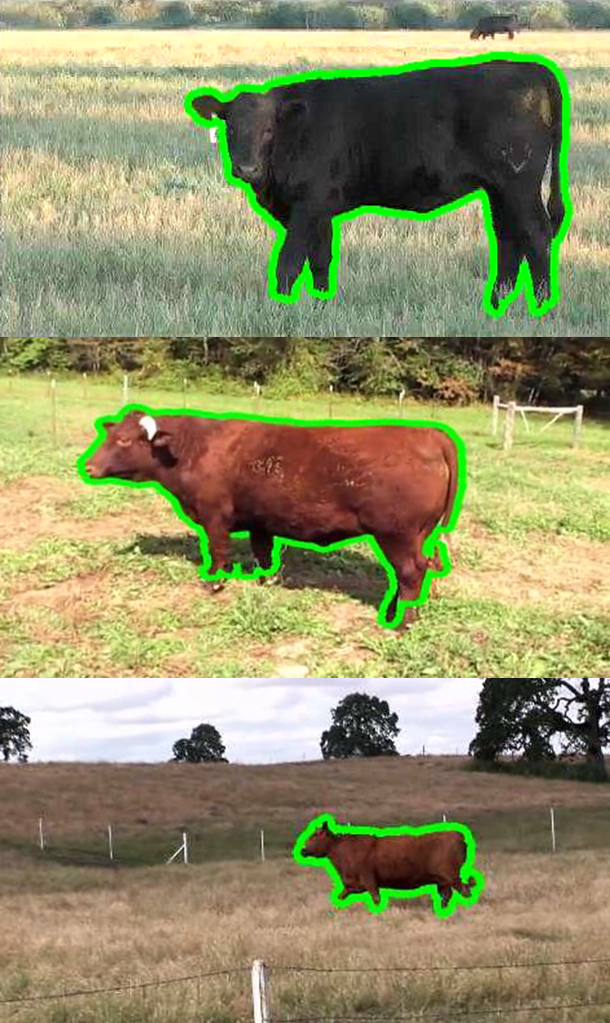}  
	}
	\subfigure[Dog]{
		\includegraphics[width=0.17\linewidth]{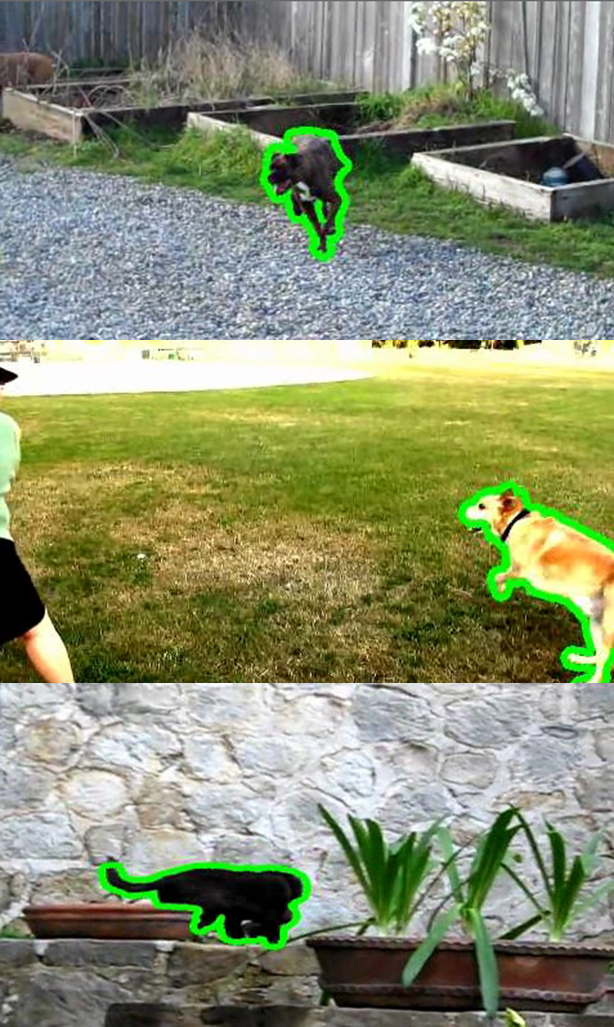}  
	}
	\subfigure[Horse]{
		\includegraphics[width=0.17\linewidth]{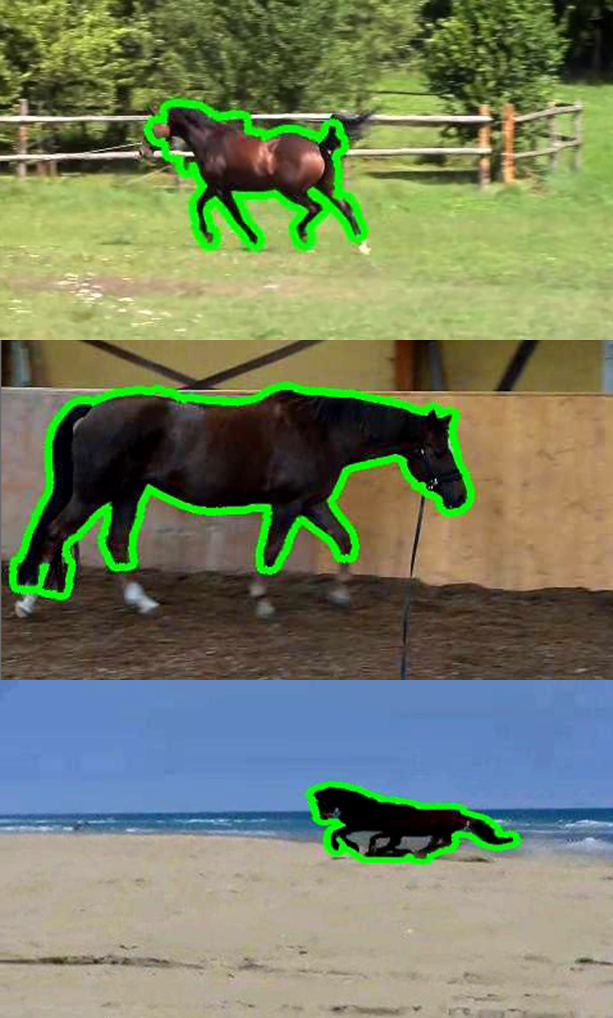}  
	}
	\subfigure[Motorbike]{
		\includegraphics[width=0.17\linewidth]{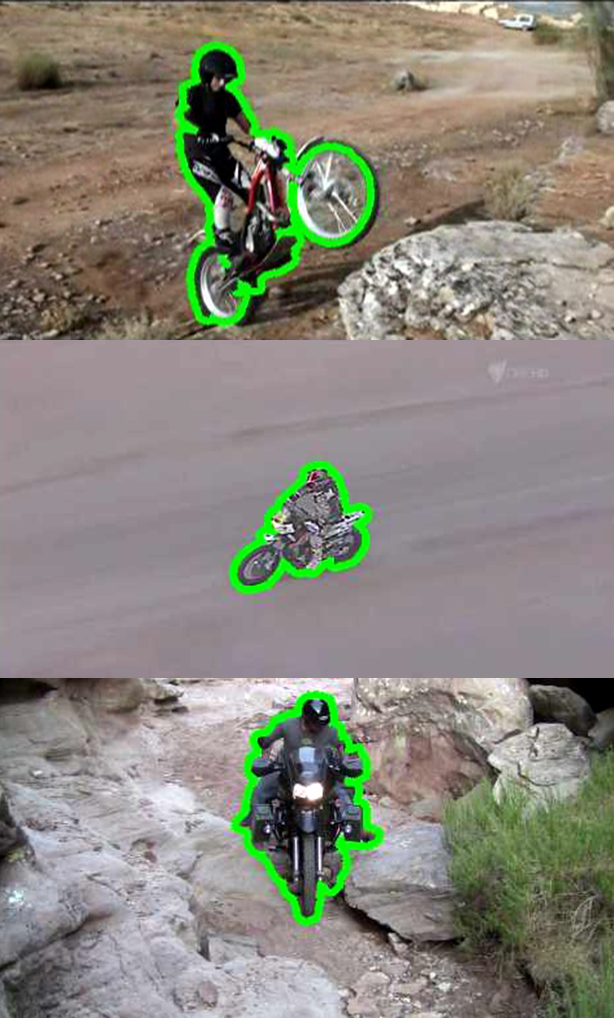}  
	}
	\subfigure[Train]{
		\includegraphics[width=0.17\linewidth]{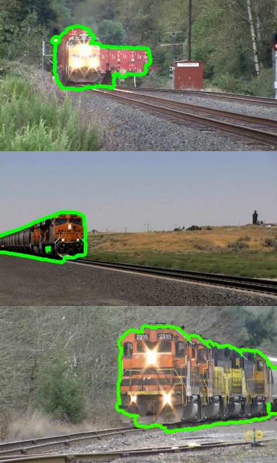}  
	}
	\caption{Representative successful results by our approach on YouTube-Objects dataset.}
\end{figure}

As ablation study, we compare the full system against two baseline systems to identify the contributions of various modules: (1) system uses confidence map from scored and classified region proposals and (2) system uses confidence map from object discovery. The first baseline scheme achieves $0.527$ in category/video average accuracy, whilst the second scheme improves the above accuracies by $3.4\%$ respectively by exploiting multiple instances via an iterative tracking and eliminating strategy. By comparing the full system with the second baseline, we observe that the proposal selection algorithm contributes $2.3\%$ respectively. 

\section{Conclusion}

We have  proposed a novel semantic object segmentation algorithm in weakly labeled video by harnessing the pre-trained image recognition model. Our core contribution is a video object proposal selection algorithm formulated as a submodular function, supported by the multiple instance learning via an iterative tracking and eliminating approach, which underpins a robust semantic segmentation method for unconstrained natural videos.

\bibliographystyle{IEEEbib}
\bibliography{refs}

\end{document}